\documentclass{article}
\usepackage{graphicx}
\usepackage{helvet}
\usepackage{authblk}
\usepackage{courier}
\usepackage[hyphens]{url}
\usepackage{graphicx}
\usepackage{amsmath} 
\usepackage{multirow}
\usepackage{booktabs}
\usepackage{mathrsfs}
\usepackage{amsfonts}
\usepackage{amssymb}  
\usepackage{array}   
\usepackage{booktabs} 
\usepackage{natbib}
\usepackage{caption}
\usepackage{algorithm}
\usepackage{algorithmic}
\usepackage{newfloat}
\usepackage{listings}
\usepackage{microtype} 
\usepackage{bm}

\date{}
\title{PEGS: Physics-Event Enhanced Large Spatiotemporal Motion Reconstruction via 3D Gaussian Splatting}

\author[1]{Yijun Xu\textsuperscript{\dag}}
\author[2]{Jingrui Zhang\textsuperscript{\dag}}
\author[1]{Hongyi Liu}
\author[3]{Yuhan Chen}
\author[2]{Yuanyang Wang}
\author[2]{Qingyao Guo}
\author[2]{Dingwen Wang}
\author[4]{Lei Yu\textsuperscript{*}}
\author[1]{Chu He\textsuperscript{*}}

\affil[1]{School of Electronic Information, Wuhan University}
\affil[2]{School of Computer Science, Wuhan University}
\affil[3]{College of Mechanical and Vehicle Engineering, Chongqing University}
\affil[4]{School of Artificial Intelligence, Wuhan University}

\begin{document}

\maketitle

\footnotetext[1]{\textsuperscript{\dag} These authors contributed equally to this work.}
\footnotetext[2]{\textsuperscript{*} Corresponding authors: Chu He (chuhe@whu.edu.cn) and Lei Yu (ly.wd@whu.edu.cn).}

\begin{abstract} \microtypesetup{protrusion=false} 
Reconstruction of rigid motion over large spatiotemporal scales remains a challenging task due to limitations in modeling paradigms, severe motion blur, and  insufficient physical consistency. In this work, we propose \textbf{PEGS}, a framework that integrates \textbf{P}hysical priors with \textbf{E}vent stream enhancement within a 3D \textbf{G}aussian \textbf{S}platting pipeline to perform deblurred target-focused modeling and motion recovery. We introduce a cohesive triple-level supervision scheme that enforces physical plausibility via an acceleration constraint, leverages event streams for high-temporal resolution guidance, and employs a Kalman regularizer to fuse multi-source observations. Furthermore, we design a motion-aware simulated annealing strategy that adaptively schedules the training process based on real-time kinematic states. We also contribute the first RGB-Event paired dataset targeting natural, fast rigid motion across diverse scenarios. Experiments show PEGS’s superior performance in reconstructing motion over large spatiotemporal scales compared to mainstream dynamic methods.
\end{abstract}

\begin{figure*}[t]
    \centering
    \includegraphics[width=\textwidth]{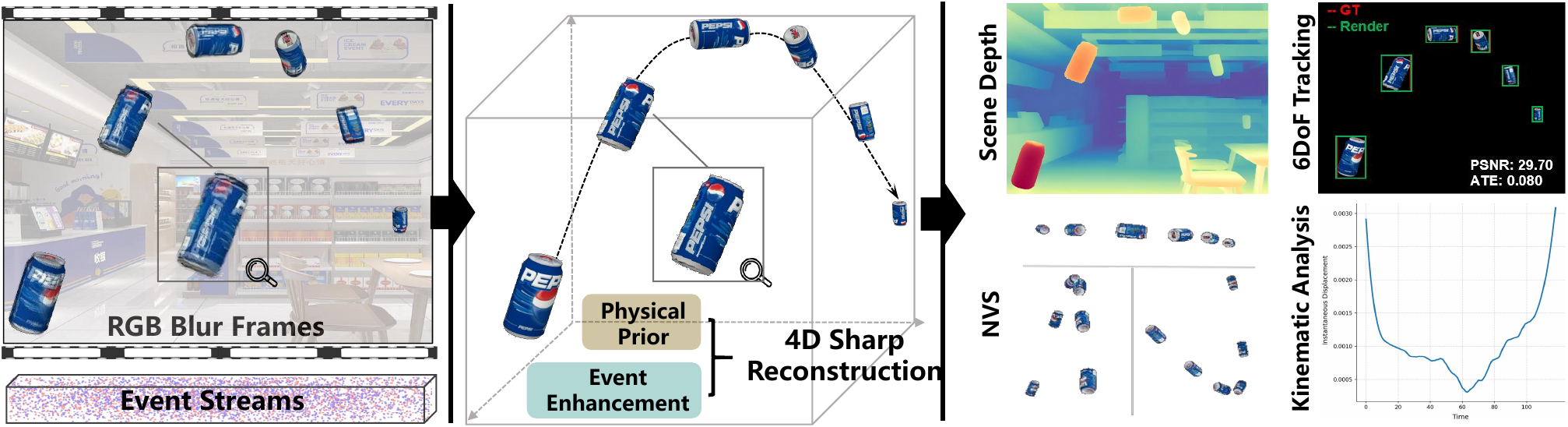} 
    \caption{PEGS takes a monocular blurry video and event streams as input to perform 4D motion recovery. The framework first focuses on the target for deblurred 3D Gaussian reconstruction, then estimates the SE-3 transformations of the motion sequence. By integrating physical priors with event enhancement, PEGS effectively reconstructs large motions, producing outputs applicable to downstream tasks. }
    \label{method}
\end{figure*}

\section{Introduction}

Dynamic reconstruction acts as the driving force of modern film, interactive gaming and virtual reality. Integrating generative approaches \cite{Dream4DGS, Dreamscene4D, PEGAUS} enables breakthroughs in controllable dynamic synthesis, pushing visual expression beyond imaginative boundaries. Concurrently, subtle deformations such as elastic collision \cite{Spring} and delicate biological tremors \cite{Endo} can be meticulously captured.

Yet, a \textit{dark cloud} still looms: \textbf{Reconstruction of complex rigid motion over long spatiotemporal spans} remains an underexplored field. The difficulties lie in: (1) modeling paradigms; (2) motion blur; (3) physical consistency.

Most existing paradigms \cite{DynaNeRF2, DyNeRF, DeformGS, 4DGS, Dynamic3DGS} are tailored for non-rigid motions that exhibit limited spatial range and temporal duration. The discrete sampling deformation framework struggles to represent intense nonlinear motion, often resulting in severe artifacts and trajectory fragmentation \cite{4DGS}. In parallel, the commonly used benchmarks \cite{DyNeRF, DynaNeRF2, DynaNeRF3} share these shortcomings: most are characterized by minimal clear deformation and are constructed under idealized assumptions. As a result, the focus has shifted towards optimizing photometric similarity, leaving the challenge of accurate motion recovery comparatively underexplored.

Motion blur stems from the inherent imaging limitations of RGB cameras. Event camera, as a bio-inspired sensor, offers a promising solution to this issue. By asynchronously recording brightness changes and generating microsecond-level event pulses, it exhibits low latency and high dynamic range. Some studies \cite{e2gs, EaDeblur, event3dgs, e4dgs, Eventboost, degs} integrate event streams into neural rendering under blur or low-light conditions. The applicability of these approaches is primarily limited to static scenes or subtle deformation blur, failing to handle challenging cases such as fast moving object.

Another challenge arises from the absence of physical consistency. Relying solely on photometric supervision struggles to constrain the vast 3D solution space, especially for motions across large spatiotemporal spans. NeRF-based implicit methods \cite{NeRFPlayer,DyNeRF,DynaNeRF3} encode motion cues within black-box network, making it difficult to model physical interactions. In contrast, 3DGS \cite{3DGS} provides structural support for scene understanding through its explicit representation. Consequently, some works \cite{Endo,CF,PEGAUS,SAGS,particlegs} incorporate off-the-shelf models to enhance physical realism. However, the model stacking essentially layers new approximations upon existing ones, thereby introducing compounded uncertainty. More critically, it bypasses the first principles in physics, ultimately reducing the reconstruction task to an uninterpretable compensatory computation.

To address these issues, we propose PEGS—a Gaussian Splatting framework that integrates physical priors with event enhancement, aiming to achieve large spatiotemporal motion reconstruction. Unlike deformation field works, we leverage the temporal continuity of video sequences and the explicit Gaussian representation to directly estimate per-frame SE-3 affine transformations. We design a cohesive triple-level supervision scheme: (1) To overcome the physical distortion in vision-only approaches, we propose an acceleration consistency constraint derived from Newtonian dynamics for precise trajectory estimation. (2) To tackle blurring caused by fast motion, we introduce event stream supervision to enhance the recovery of dynamic details. (3) Furthermore, we devise a Kalman regularizer to unify multi-source observations and minimize estimation errors. 

Overall, our main contributions can be outlined as:

\begin{itemize}

\item We introduce PEGS, a framework for large spatiotemporal motion reconstruction. Particularly, we contribute the first RGB-Event paired dataset targeting natural motion across large spans with diverse scenarios.

\item We developed a triple-level supervision scheme with acceleration constraints, event enhancement, and Kalman regularization to tackle the critical issues of physical absence, motion blur, and multi-source observation.

\item We developed a motion-aware simulated annealing (MSA) strategy that adaptively schedules the training process based on real-time velocity and displacement, thereby enhancing convergence efficiency and stability.

\end{itemize}

\section{Related Work}
\label{sec:related}

\subsection{Dynamic Neural Rendering}
Current dynamic neural rendering can be categorized into: (1) Point deformation \cite{DynaNeRF2}; (2) Time-aware volume rendering \cite{DyNeRF, NeRFPlayer}; (3) Gaussian deformation\cite{DeformGS, 4DGS, MotionGS, S4d}. A common issue lies in the difficulty of deformation fields in modeling large inter-frame displacements. Furthermore, the networks prioritize fitting appearance similarity while neglecting geometric accuracy. A typical manifestation is that when large areas of approximately uniform color exist, Gaussian splatting tend to undergo disordered drift within these color-homogeneous regions \cite{Dynamic3DGS, CF}.

\subsection{Physics-Enhanced Dynamic 3DGS}
Integrating physical assistance to enhance 3DGS has gained widespread adoption. Works such as \cite{CF, Endo, FSGS} leverage off-the-shelf monocular depth estimation \cite{DPT} for scene initialization or sparse reconstruction. \cite{Spring} devises a Spring-Mass model to simulate the falling and collision behaviors of elastic objects. Tailored for 3D dataset generation, \cite{PEGAUS} introduces the PyBullet \cite{pybullet} to emulate object's placement. \cite{Dreamscene4D, SAGS} improve pose estimation by enforcing constraints on scale consistency and local rigidity. Current approaches largely rely on off-the-shelf models to alleviate the ill-posed problems inherent in vision-only systems. However, pre-trained engines exhibit a sharp drop in confidence when confronted with unusual scenarios, and their black-box nature renders errors untraceable. Differently, we propose a novel strategy that is directly grounded in the first principles of physics \cite{particlegs}, rather than depending on external models.

\subsection{Event-based Neural Reconstruction}
Event-based neural rendering has emerged as a promising direction for 3D scene perception. Early methods \cite{e2nerf, evdeblurnerf} combined events with RGB to recover clear 3D scenes from blurred inputs, and \cite{denerf} extended this to dynamic settings. Recently, focus has shifted to integrating events into 3DGS \cite{3DGS}, leading to methods like \cite{e2gs, EaDeblur, event3dgs} for static scenes, and \cite{e4dgs, Eventboost, degs} for dynamic ones. Nevertheless, these approaches are still constrained to small camera motions or minor object deformations. Another bottleneck lies in dataset collection. Most studies are verified on synthetic RGB-E pairs \cite{e4dgs, degs}, with images rendered in Blender \cite{Blender} and events simulated via V2E \cite{v2e}. Though some efforts have been made to collect real-world data \cite{Darkgs, Eventboost}, they are typically captured using controlled turntables, resulting in limited scene complexity and motion patterns. In contrast, we target a more challenging setting—deblurring under large span, fast motion, and introduce a data acquisition scheme detailed in \ref{exper}.

\section{Preliminaries}
\label{sec:pre}

\subsection{3D Gaussian Splatting}
As a point-based representation, 3DGS \cite{3DGS} structures a scene into a set of Gaussian ellipsoids, each characterized by the center location $x$ and a 3D covariance matrix $\bm{\Sigma}$: 
\begin{equation}
\bm{\mathcal{G}}(\bm{x}) = exp(-\frac{1}{2}\bm{x}^{T}\Sigma^{-1}{\bm{x}}),
\label{1_GS}
\end{equation}
where the $\bm{\Sigma} = \bm{RSS}^T\bm{R}^T$ is parameterized by scaling matrix $\bm{S}$ and rotation matrix $\bm{R}$.

During the splatting, Gaussians are projected onto a plane. This relies the 2D  covariance matrix $\bm\Sigma' = \bm{JW \Sigma W}^T\bm{J}^T$ in the camera coordinate, where $\bm{W}$ is the view transformation matrix, and $\bm{J}$ is the Jacobian matrix of the affine projection. For rendering, the color $\bm{C}$ of each pixel is determined by the combined effect of the corresponding $N$ depth-ordered Gaussians' color $\bm{c_i}$ and opacity $\alpha_i$:
\begin{equation} 
\bm{C} = \sum_{i \in N} \bm{c}_i \alpha_i \prod_{j=1}^{i-1} (1 - \alpha_j).
\label{2_render}
\end{equation}

\subsection{Event Signals}
\label{EDI}
When an event sensor detects that the brightness change at pixel $(x, y)$ exceeds the threshold $\epsilon$, it asynchronously reports an event polarity $p$:
\begin{equation}
p = \begin{cases} 
+1 & \text{if } log(x, y, \tau) - log(x, y, \tau') > \epsilon \\
-1 & \text{if } log(x, y, \tau') - log(x, y, \tau) > \epsilon,
\end{cases}
\label{3_event}
\end{equation}
where $\tau_0$ and $\tau$ denotes the first and the last timestamp of the observed event respectively. Introducing a unit integral impulse function $\delta(\tau)$, the discrete polarities can be expressed as a continuous event signal: $\bm{e}(\tau) = p\delta(\tau)$. Based on this, the proportional intensity change over the time interval can be computed as the integral of the event stream:
\begin{equation}
\bm{E}(\tau) = \int_{\tau_0}^{\tau} e(\tau)d\tau,
\label{4_event_intergral}
\end{equation}
then the image $\bm{I}_{\tau}$ captured at time $\tau$ can be computed as:
\begin{equation}
\bm{I}(\tau) = \bm{I}(\tau_0)\cdot exp(\epsilon\bm{E}(\tau)).
\label{eq.event_intergral}
\end{equation}

According to the assumption of the Event-based Double Integral (EDI) model \cite{EDI}: A blurry image is the average of $N$ latent sharp images over the exposure time. Therefore, given the threshold $\epsilon$, the blurry image $\bm{I}_{\tau}$, and the corresponding event stream $\bm{E}(\tau)$, latent sharp images at any instant during the exposure time can be recovered.

\section{Method}
\label{sec:method}

\begin{figure*}[t]
    \centering
    \includegraphics[width=\textwidth]{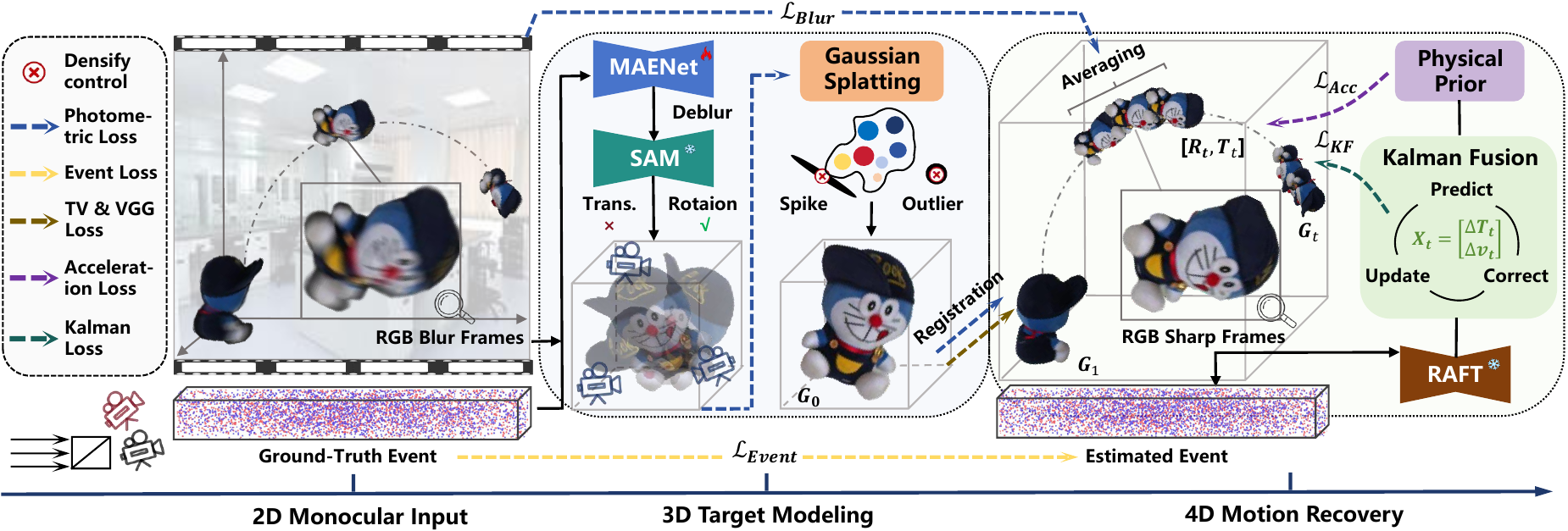} 
    \caption{PEGS reconstructs a target-focused 3D Gaussian scene from blurry images and then estimates its full SE(3) motion trajectory. This is achieved by combining event-based deblurring with a triple-level motion supervision strategy that enforces acceleration consistency, event stream alignment, and Kalman regularization. A motion-aware simulated annealing scheduler further boosts training convergence.}
    \label{method}
\end{figure*}

The PEGS framework consists of two tightly coupled stages: target modeling and motion recovery (see Fig.\ref{method}). The first stage leverages event-based deblurring and an improved point density control strategy to reconstruct a target-focused 3D Gaussian representation from blurry RGB inputs. The subsequent motion recovery stage estimates the complete SE-3 motion trajectory through a triple-level supervision strategy comprising: an acceleration consistency constraint derived from Newtonian dynamics, high-temporal resolution event stream supervision, and Kalman regularization that fuses multi-source observations. Furthermore, we introduce a MSA strategy that adaptively schedules the training process to enhance convergence efficiency and stability.

\subsection{Target Modeling} 
The task of this stage is to reconstruct target-focused Gaussians from blurred input. The pipeline comprises event-based deblurring, target centering, GS reconstruction incorporating an improved point density control strategy, and finally, inversely registering the Gaussians from the centered coordinate back to the original complete scene.

\textbf{Image Deblurring.} The preliminary workflow of 3DGS \cite{3DGS} involves initial point cloud and camera pose estimation, which typically requires COLMAP package \cite{SfM}. Unfortunately, this process often fails for blurry image sequences \cite{e2nerf, e2gs}. For image deblurring, based on EDI model \cite{EDI} and MAENet \cite{MotionAware}, we divide the exposure duration corresponding to each blurry image into $N$ timestamps, thereby obtaining $N-1$ corresponding event bins $\{\bm{B}_i\}_{i=1}^{N-1}$, and subsequently recover $N$ deblurred images $\{\bm{I}_i\}_{i=1}^{N}$. 

\textbf{Centralization.} Based on the deblurred image, we first employ the SAM \cite{SAM} to separate dynamic objects from static backgrounds. Subsequently, by establishing an object-centric normalized coordinate \cite{Dreamscene4D}, we achieve the decomposition of complex projectile motion: the translational component is eliminated, and the object's autorotation is equivalent to pseudo multi-view observations, which are then fed into \cite{SfM} for initializing $N$ pesudo camera poses $\{\bm{P}_i\}_{i=1}^{N}$. The process mitigates the negative impact of complex motion on the accuracy of pseudo-view estimation, and facilitates subsequent centroid computation for physical constraints.

\textbf{Reconstruction.} Following the 3DGS \cite{3DGS}, we use a set of Gaussian kernels to explicitly reconstruct the object. Throughout the optimization, we design the loss function with reference to \cite{e2gs, evagaussians, 3DGS}. Specifically, we average $N$ rendered deblurred images $\{\bm{\hat{I}}_i\}_{i=1}^{N}$ for each corresponding blurry input, thus obtaining the rendered blurry image $\bm{\hat{I}}_{blur}$:
\begin{equation}
\bm{\hat{I}}_{blur} = \frac{1}{N}\sum_{i=1}^{N} \bm{\hat{I}}_i,
\label{eq.deblur}
\end{equation}
then the real blurry input $\bm{I}_{blur}$ is used as the supervisory signal to compute the loss:
\begin{equation}
\mathcal{L}_{blur} = (1-\lambda)\lVert{\bm{I}_{blur}-\bm{\hat{I}}_{blur}}\rVert_{1} + \lambda\mathcal{L}_{D-SSIM}.
\label{blurloss}
\end{equation} 

Notably, the original 3DGS \cite{3DGS} framework exhibits limitations in geometric accuracy when modeling focal objects, where ill-defined boundaries adversely affect subsequent centroid localization. To address this, we implement an improved point density control strategy \cite{pmgs} to hard prune the jagged edges and spiky artifacts, while eliminating isolated points that float distant from the target.

Up to this point, we have reconstructed $\bm{\mathcal{G}}_0$ at a centralized scale with refined appearance and geometric representation from blurry inputs.

\textbf{Registration.} Before performing motion trajectory recovery, we register $\bm{\mathcal{G}}_0$ under the centered scale to the original complete dynamic scene. We use a set of learnable affine transformations  $\bm{T}_{reg} = [\bm{r, t}, s]$ to achieve this, i.e., quaternion rotation $\bm{r}\in \mathbb{R}^{4}$, translation vector $\bm{t}\in \mathbb{R}^{3}$, and constant scale $s$. Thus the target Gaussian $\bm{\mathcal{G}}_1$ can be represented as $\bm{T}_{reg}\odot \bm{\mathcal{G}}_0$, and the optimization is to minimize $\mathcal{L}_{Reg}$ between the rendered image of $\bm{\mathcal{G}}_1$ and the first deblurred frame $\bm{{I}}_0$ of original dynamic scene:
\begin{equation}
\bm{T}_{reg} = \mathrm{arg}\underset{\bm{T}_{reg}}{\mathrm{min}}\mathcal{L}_{reg}(Render(\bm{\mathcal{G}_1}), \bm{I}_0).
\label{Equation6}
\end{equation}

In addition to $\mathcal{L}_{GS}$ \cite{3DGS}, we further apply VGG loss $\mathcal{L}_{VGG}$ and grid-based total variation loss $\mathcal{L}_{TV}$ \cite{Spring, 4DGS} to achieve a better awareness for accurate registration. The total registration loss can be formulated as:
\begin{equation}
\mathcal{L}_{reg} = \lambda_{1}\mathcal{L}_{GS}+ \lambda_{2}\mathcal{L}_{VGG} + \lambda_{3}\mathcal{L}_{TV}.
\label{Equation7}
\end{equation}

During the entire modeling phase, we reconstruct a clear Gaussians $\bm{\mathcal{G}}_1$ of the target from blurry inputs. We will proceed to recover the full motion sequence based on this.

\subsection{Motion Recovery}
This stage recovers the full motion sequence through frame-wise estimation of the object's SE(3) pose transformations. As mentioned before, in a video sequence, each blurry input at time $t$ corresponds to $N$ deblurred images with a constant exposure interval $\Delta t$, which yields $N-1$ pose changes $\bm{\hat{P}}_t = [\bm{\hat{R}}_t, \bm{\hat{T}}_t]$ to be learned. 

In addition to the fundamental image-level blur loss (consistent with Eq.\ref{blurloss}, but applying the complete non-centralized blurry input as ground-truth), we introduce a cohesive triple-level supervision scheme: (1) \textit{Physics-level Acceleration Loss} to enforce temporal smoothness and physical consistency; (2) \textit{Event-level Loss} to supervise high temporal resolution motion details; and (3) \textit{Fusion-level Kalman Loss} that estimates the optimal pose from multi-source observation as a regularization term. Details are as follows:

\textbf{Acceleration Consistency Loss.} An object moving in a constant force field maintains constant acceleration. Based on this, we impose smoothness constraints on the acceleration to ensure the estimated displacements adhere to physical consistency. Firstly, we calculate the object's center of mass. Since we have implemented point density control and isotropic processing, the centroid can be approximated as:
\begin{equation}
\bm{\sigma} = \frac{\sum_{i=1}^MR_i^3\bm{\mu}_i}{\sum_{i=1}^MR_i^3},
\label{Equation8}
\end{equation}
where $R_i$ and $\bm{\mu}_i$ denote the radius and position of total $M$ kernels. In turn, the object's acceleration at timestamp $t$ can be calculated from the displacement of centroid as:
\begin{equation}
\bm{a}_{t}=\frac{[\bm{\sigma}_{(t+\Delta{t})}-\bm{\sigma}_t]-[\bm{\sigma}_t-\bm{\sigma}_{(t-\Delta{t})}]}{(\Delta t)^2},
\label{acc}
\end{equation}
where $\Delta{t}$ denotes the time interval between two adjacent deblurred frames. Based on this, we minimize the acceleration variation between consecutive timestamps:
\begin{equation}
\mathcal{L}_{Acc} = \sum_{t=0}^{T}\lVert{\bm{a}_{t} 
 - \bm{a}_{t+\Delta t}}\rVert_{2}^{2}.
\label{lossaccneed}
\end{equation}

\textbf{Event Stream Loss.} Leveraging the microsecond-level resolution of event cameras, we further impose constraints on the continuous motion recovery. As mentioned earlier, each blurred frame corresponds to $N-1$ event bins $\{\bm{B}_i\}_{i=1}^{N-1}$ and $N$ deblurred frames $\{\bm{I}_i\}_{i=1}^{N}$. According to \cite{EDI}, we integrate these to obtain $N-1$ event maps $\{\bm{E}_i\}_{i=1}^{N-1}$, which serve as the ground truth supervision signals. Correspondingly, we synthesize  $N-1$ simulated event sequences $\{\bm{\hat{E}}_i\}_{i=1}^{N-1}$ by computing the intensity changes between consecutive $N$ rendered images \cite{e2gs}. The estimation of motion details is thereby constrained by comparing the differences between the simulated and real event maps:
\begin{equation}
\mathcal{L}_{Event} = \sum_{i=1}^{N-1}\lVert{\bm{E}_{i} 
 - \bm{\hat{E}}_{i}}\rVert_{1}.
\label{lossacc}
\end{equation}

\textbf{Kalman Fusion Loss.}
Though event signals provide sufficient motion cues, the data is inherently sparse with unavoidable noise. On the other hand, dynamic models offer strong constraints on object motion, but real-world disturbances such as air resistance and non-ideal force fields can cause deviations in physical predictions. To this end, we employ a Kalman Fusion (KF) \cite{Kalman} to combine the physical model with event stream observations, thereby obtaining an optimal estimation reference. Detailed steps are as follows:

\noindent
\textbf{Step1 - Initialization:} We begin by mathematically modeling the KF process:

\begin{itemize}
\item {\textit{State vector.}} We define the object's instantaneous displacement and velocity as the state vector $\bm{X}_t = [\bm{T}_t, \bm{v}_t]$.

\item {\textit{System model.}} Based on acceleration constraints (described by Eq.\ref{lossaccneed}), the displacement at current moment can be predicted from the previous optimal estimate.

\item {\textit{Observation.}} We feed the clear images deblurred by events into RAFT \cite{eraft} to compute the optical flow, from which the observed instantaneous displacement $\bm{\tilde{T}}_{t}$ is mapped.
\end{itemize} 

\noindent
\textbf{Step2 - Prediction:} Based on the system model and the optimal estimate $\bm{X}_{t|t-\Delta t}$ from the previous moment, we predict the current state vector $\bm{X}_{t}$:
\begin{equation}
\begin{aligned}
& \bm{X}_{t|t-\Delta t} = \bm{F}\bm{X}_{t-\Delta t} + \bm{Ga}_t + \bm{w}_{t}, \\
& \bm{F} = \begin{bmatrix}
  \bm{I} & \Delta t \\
  0 & \bm{I} 
  \end{bmatrix}, \bm{G} = \begin{bmatrix}
   \frac{1}{2}(\Delta t)^2\\
  \Delta t \end{bmatrix}, \bm{Q} =\begin{bmatrix}
  \sigma_{\bm{T}}^{2} & 0 \\
  0 & \sigma_{\bm{v}}^{2} 
  \end{bmatrix}.
\label{predict}
\end{aligned}
\end{equation}

The state transition is governed by $\bm{F}$. $\bm{G}$ denotes the control input matrix, and the control term $\bm{a}_t$ is computed from Eq.\ref{acc}. Process noise $\bm{w}_t$ (comprising $\sigma_{\bm{T}}^{2}$ and $\sigma_{\bm{v}}^{2}$), which characterizes model uncertainty, has its covariance matrix defined as $\bm{Q} = \mathbb{E}[\bm{w}_{t} \bm{w}_{t}^{T}] $. The predicted covariance matrix is then obtained through the following relation:
\begin{equation}
\begin{aligned}
& \bm{C}_{t|t-\Delta t} = \bm{F}\bm{C}_{t-\Delta t} \bm{F}^{T} + \bm{Q}.
\label{P}
\end{aligned}
\end{equation}

\begin{figure}[t!]
    \centering
    \includegraphics[width=\linewidth]{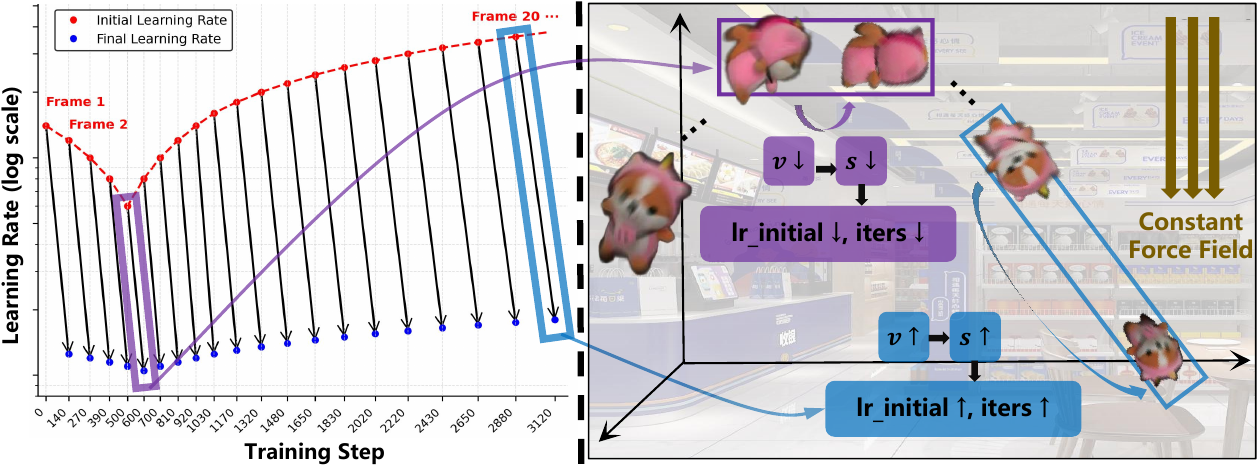} 
    \caption{MSA strategy. Bottom: An object accelerates in a constant force field, with the velocity $\bm{v}$ and displacement $\bm{s}$ vary at each timestamp. Top: The red curve traces the initial learning rate, and blue points signify its value after exponential decay.}
    \label{MSA}
\end{figure} 

\noindent
\textbf{Step3 - Correction:} In the update step, we introduce observation to correct the predicted state: $ \bm{z}_{t} = \bm{H}\bm{\tilde{T}}_{t} + \bm{u}_{t}$,
where $\bm{H} = [\bm{I} ,0]$ is the observation matrix, $\bm{R} = \mathbb{E}[\bm{u}_{t}\bm{u}_{t}^{T}] $ is the covariance of event observation noise.

\noindent
\textbf{Step4 - Updation:}
Based on prediction and correction, we can dynamically update the Kalman gain $\bm{K}_{t}$ and the posterior covariance $\bm{C}_{t}$:
\begin{equation}
\bm{K}_{t} = \bm{C}_{t|t-\Delta t}\bm{H}^{T}(\bm{H}\bm{C}_{t|t-\Delta t}\bm{H}^{T}+\bm{R})^{-1},
\end{equation}
\begin{equation}
\bm{C}_{t} = (\bm{I}-\bm{K}_{t}\bm{H})\bm{C}_{t|t-\Delta t}.
\end{equation} 

Finally, we obtain the optimal estimated value at the current timestamp:
\begin{equation}
\bm{X}_{t} = \bm{X}_{t|t-\Delta t}+ \bm{K}_{t}(\bm{z}_{t}-\bm{H}\bm{X}_{t|t-\Delta t}).
\end{equation}

Therefore, the translation component $\bm{T}_{t}$ from the optimal estimate $\bm{X}_t$ will be updated to the reference signal $\bm{P}_t$ to supervise the learning of the target $\bm{\hat{P}}_t$.
\begin{equation}
\mathcal{L}_{KF} = \sum_{t=0}^{T}\lVert{\bm{P}_{t} 
 - \bm{\hat{P}}_{t}}\rVert_{2}^2.
\label{lossacc}
\end{equation}

Essentially, the KF process performs Bayesian estimation of event observations with a Newtonian mechanics model, thereby providing an adaptively calibrated regularization with multi-source observation fusion for training.

In summary, the total loss for motion recovery stage is:
\begin{equation}
\mathcal{L}_{Total} = \lambda_1 \mathcal{L}_{Blur} + \lambda_2 \mathcal{L}_{Acc} + \lambda_3 \mathcal{L}_{Event} + \lambda_4 \mathcal{L}_{KF}.
\label{lossacc}
\end{equation}

\subsection{Motion-aware Simulated Annealing}
Fixed optimization paradigms struggle to accurately model variable-speed motion. Specifically, a fixed initial learning rate $l_{init}$ may fail during high-speed motion phases, while exhibiting significant oscillations at low speeds.

To address this, we design a novel MSA strategy, as shown in Fig.\ref{MSA}. Based on the optimal estimation of instantaneous displacement $\bm{T}_{t}$ from KF process, we adaptively schedule the learning rate of the next timestamp:
\begin{equation}
l(n)_{t+\Delta t} = (\frac{l_{max}-l_{min}}{|\bm{T}_{max}|}\cdot|\bm{T}_{t}|+ l_{min})\cdot\gamma^{e(n)},
\label{lossacc}
\end{equation}
where $|\bm{T}_{max}|$ represents the maximum observed displacement, $l_{min}$ and $l_{max}$ are the empirically determined values. This implies that large displacements require a large learning rate for fast convergence, whereas small displacements need a small learning rate to avoid oscillations.

As the step index $n$ increases during training, the Gaussians progressively converge toward target spatial positions. Thus we implement a gradually shrinking step size via exponentially decaying $e(n)$ to enable refinement, which controlled by the decay factor $\gamma \in (0,1)$.

\begin{table}[t]
\centering
\renewcommand{\arraystretch}{1}
\setlength{\tabcolsep}{5pt}

\resizebox{\textwidth}{!}{
\begin{tabular}{l c c c c c c}
\hline
Datasets 
& Source & Phy. & Mono. & Mod.  & Type\\
\hline

D-NeRF \cite{DynaNeRF2}    
& S   & $\times$ & $\checkmark$ & RGB & Def. \\

Neural3D  \cite{DyNeRF} 

& S  & $\times$ & $\times$ & RGB & Def.\\

PEGASUS \cite{PEGAUS}
& S  & PyBullet & $\checkmark$  & RGB & Rigid\\

SpringGS \cite{Spring}   
& 6S + 3R &  Hooke's Law &  $\times$  & RGB  & Def.\\

DE-NeRF \cite{denerf}
& 3S + 3R   & $\times$ & $\times$ & RGB+E & Def.\\

E-4DGS \cite{e4dgs}
& 8S + 3R   & $\times$ & $\times$ & RGB+E & Def. \\

Event-boost \cite{e-boost}
& 8S + 4R   & $\times$ & $\times$ & RGB+E & Def. \\

PEGS 
& 6S + 6R   & Newton's Law & $\checkmark$ & RGB+E & Rigid\\

\hline
\end{tabular}
    }
\caption{Datasets comparison (Phy.-Whether physical prior is introduced; Mono.-Whether monocular; Mod.-Modality; S-Synthetic and R-Real; Def.-Deformable).}
\label{datatable}
\end{table}

\begin{figure}[t!]
\centering
\includegraphics[width=\linewidth]{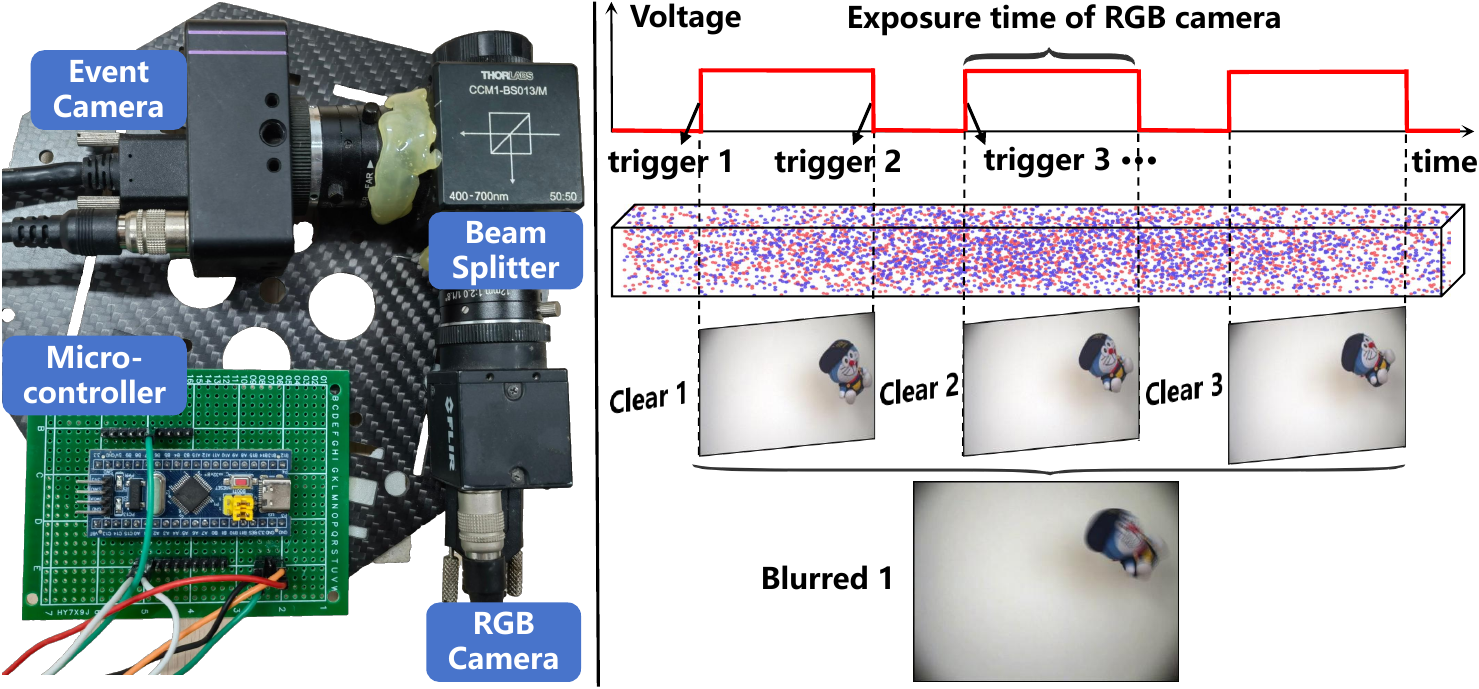}
\caption{Schematic of RGB-Event spatiotemporal synchronization acquisition device (left) and imaging process (right).}
\label{data}
\end{figure}

\section{Experiments}
\subsection{Experimental Settings}\label{exper}
\textbf{Datasets.} 
By extending the primary part of the dataset established by \cite{pmgs}, we constructed the first dataset dedicated to natural rigid motion across large spatiotemporal scales. It covers diverse fundamental dynamics: horizontal / oblique projectile motion, free fall and friction. The dataset consists of 6 sets of synthetic and 9 sets of real-world data. Table.\ref{datatable} summarizes the comparisons with various benchmarks.

\begin{itemize}
\item Synthetic scenes. We collected 6 models from the public 3D community \cite{Blender}, projecting them with random initial velocities and autorotation. A uniform force field was deployed to simulate real conditions (e.g., gravity and resistance), which is rarely considered in previous benchmarks. A fixed camera was used to capture at 120 FPS and 1024*1024 resolution. Clear images were averaged after superimposition to synthesize blurred images, and event streams were simulated by V2E \cite{v2e}.

\item Real-world scenes. The field currently suffers from a severe scarcity of 4D RGB-E datasets, with very few being open-sourced \cite{e4dgs, e-boost}. The only publicly available dataset \cite{denerf} remains limited to minor deformations. Furthermore, unlike common ideal conditions \cite{e-boost, Darkgs}, where objects are placed on a controlled turntable rotating uniformly, we built an RGB-Event spatiotemporally synchronized imaging device to flexibly capture scenes, as shown in Fig.\ref{data}. The setup uses an STM32 microcontroller to trigger PWM signals controlling a FLIR camera for imaging (720*540). Light is spatially synchronized to a DVXplorer camera (640*480) via a beam splitter to record event streams.

\end{itemize}

In the novel view reconstruction task, the training and validation sets are divided according to \cite{3DGS}. For motion recovery, we focus on recovering the motion across the entire continuous time sequence (i.e., all frames), thus no split is applied. All competitors adopt the aforementioned division scheme to ensure fair metric computation.

\begin{figure}[t!]
    \centering
    \includegraphics[width=\textwidth]{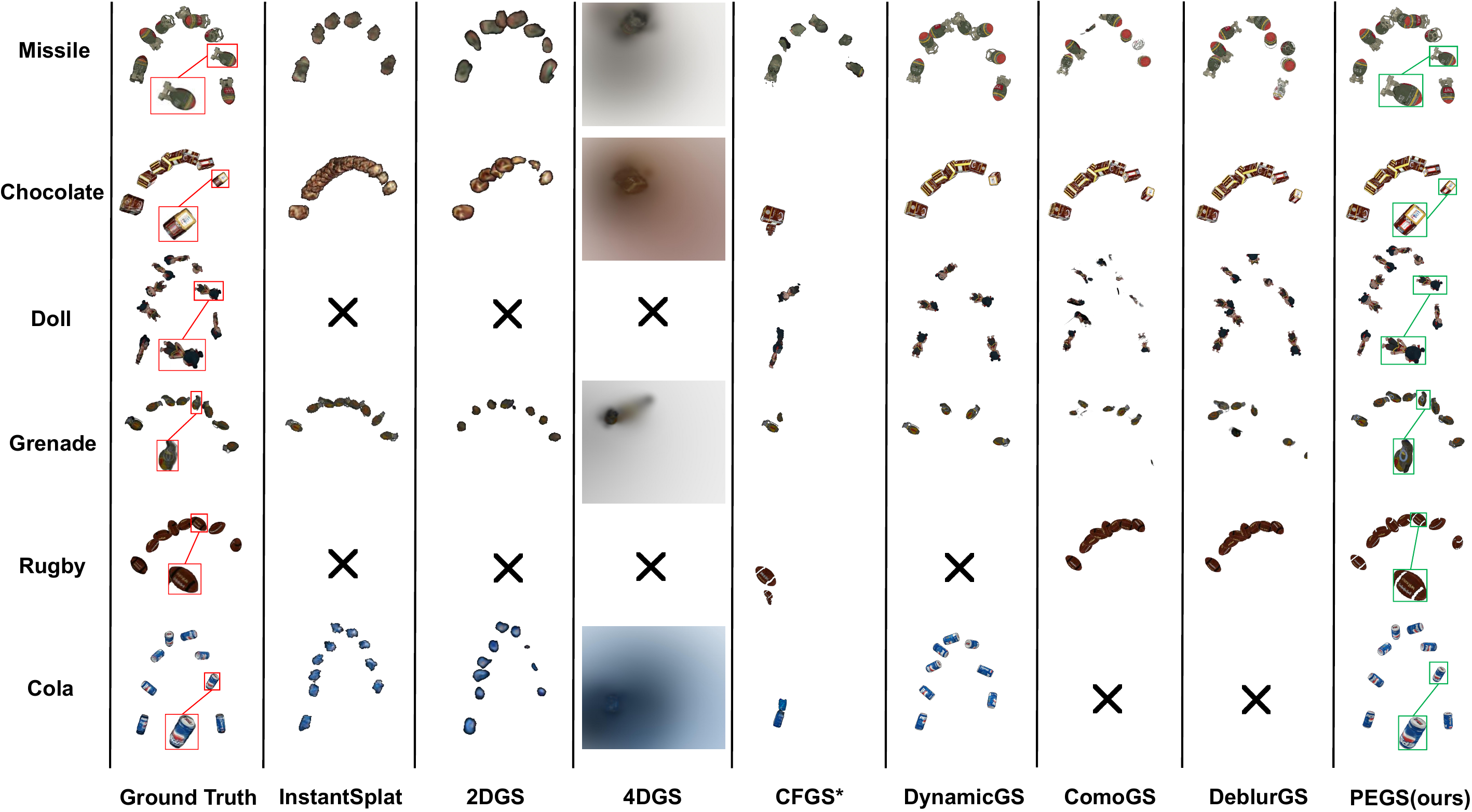} 
    \caption{Qualitative comparison on the synthetic dataset. For clearer results, please refer to the video in the supplementary material.}
    \label{comparison}
\end{figure}

\begin{figure}[t!]
    \centering
    \includegraphics[width=\textwidth]{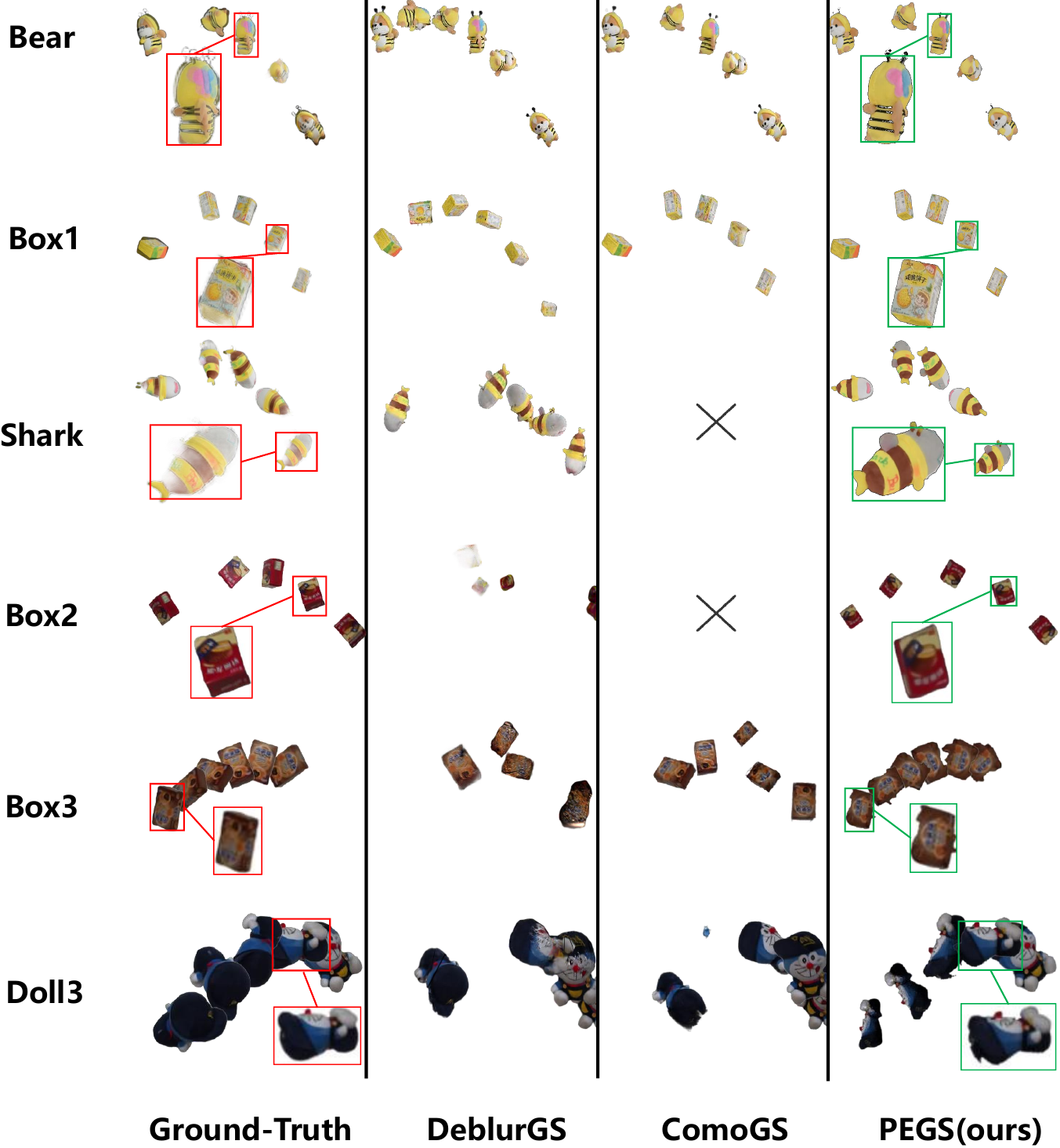} 
    \caption{Qualitative comparison on the real dataset.}
    \label{realcomparison}
\end{figure}

\textbf{Metrics.}
For \textbf{video reconstruction}, we adopt PSNR, SSIM, and LPIPS \cite{SSIM, PSNR}. For \textbf{motion recovery}, we calculate the bounding boxes of the target in both real and rendered images, and then assess the spatial accuracy of the trajectory using IoU, absolute trajectory error (ATE), and RMSE \cite{IoU, Tracking, CF}. All evaluations are performed after complete background removal.

\textbf{Baselines.} Given the severe scarcity of open-source, comparable 4D event-based algorithms, we selected RGB-only methods 4DGS \cite{4DGS}, DynamicGS \cite{Dynamic3DGS} and MotionGS \cite{MotionGS}. For a fair comparison, we fed the deblurred images from our trained MAE network into them so as to assess the baselines' dynamic modeling capability with clear input (denoted as "E + 4DGS", etc.). Baselines E2GS \cite{e2gs}, ComoGS \cite{comogs}, and DeblurGS \cite{DeblurGS} are specifically designed for blurry scene reconstruction. CFGS \cite{CF} shares a similar 6DoF pose estimation with our approach. However, its depth-based initialization fails entirely for focused targets. Therefore, we initialized it with our pre-trained Gaussians for an equitable comparison. Furthermore, considering the characteristics of rigid motion, we also included 2DGS \cite{2DGS}, and InstantSplat \cite{Instantsplat} as static-scene baselines.

\textbf{Implementation Details.} We conducted experiments using PyTorch on a NVIDIA RTX 4090 GPU. During the \textbf{target modeling} stage, we cropped the target region from the raw data and centered it on a 512×512 pixel canvas. Besides, we optimized a set of $[\mathbf{r, t}, s]$ within 10,000 iterations with a learning rate of 5e-5 to align the target-centered Gaussians with the first frame of the dynamic scene. This process is efficient and typically completes in minutes. For the \textbf{motion recovery} stage, the initial learning rate was set to 1.0e-3 for the frame with minimal velocity, with a base iteration count of 1,000. The loss weights for $\mathcal{\bm{L}}_{Blur}$, $\mathcal{\bm{L}}_{Acc}$, $\mathcal{\bm{L}}_{Event}$, and $\mathcal{\bm{L}}_{KF}$ were set to 0.7, 0.1, 0.15, and 0.05, respectively.

\begin{table}[t]
    \centering

    \renewcommand{\arraystretch}{1}
    \setlength{\tabcolsep}{3pt}
    \resizebox{1\textwidth}{!}{
    \begin{tabular}{l|*{6}{ccc|}ccc}
        \toprule
        & \multicolumn{3}{c|}{Grenade} 
        & \multicolumn{3}{c|}{Missile} 
        & \multicolumn{3}{c|}{Cola} 
        & \multicolumn{3}{c|}{Doll}
        & \multicolumn{3}{c|}{Chocolate} 
        & \multicolumn{3}{c|}{Rugby} 
        & \multicolumn{3}{c}{\textbf{Mean}}\\
        
        \cmidrule(lr){2-4}
        \cmidrule(lr){5-7} 
        \cmidrule(lr){8-10} 
        \cmidrule(lr){11-13} 
        \cmidrule(lr){14-16} 
        \cmidrule(lr){17-19}
        \cmidrule(lr){20-22} 
        
        & PSNR$\uparrow$ & SSIM$\uparrow$ & LPIPS$\downarrow$ 
        & PSNR$\uparrow$ & SSIM$\uparrow$ & LPIPS$\downarrow$
        & PSNR$\uparrow$ & SSIM$\uparrow$ & LPIPS$\downarrow$ 
        & PSNR$\uparrow$ & SSIM$\uparrow$ & LPIPS$\downarrow$ 
        & PSNR$\uparrow$ & SSIM$\uparrow$ & LPIPS$\downarrow$ 
        & PSNR$\uparrow$ & SSIM$\uparrow$ & LPIPS$\downarrow$ 
        & PSNR$\uparrow$ & SSIM$\uparrow$ & LPIPS$\downarrow$ \\
        \midrule

        E + 2DGS \cite{2DGS}
        &24.82  &0.895  &0.153
        &17.42  &0.714  &0.325
        &15.97  &0.847  &0.195
        &17.99  &0.792 &0.218
        &15.19  &0.758 &0.275
        &21.56  &0.830 &0.185
        
        &18.83   &0.806  &0.225  \\
        
        E + InstantSplat \cite{Instantsplat}
        &25.44 &0.900  & 0.139
        &17.58  &0.724  &   0.300
        &17.03   &0.852   &  0.175
        &18.10  &0.797 &  0.221
        &16.10 &0.761 & 0.255
        &24.21 &0.877  & 0.165
        
        &19.74   &0.818  &0.209 \\
        
        CFGS \cite{CF}
        & 21.82 & 0.873 & 0.231
        & 13.70 & 0.675 & 0.259
        & 14.47 & 0.845 &  0.284
        & 18.00 & 0.788 & 0.307
        & 11.93  & 0.739  &  0.317
        &21.37 &0.822 & 0.196
        
        &16.88   &0.790  &0.266 \\

        E + 4DGS \cite{4DGS}
        & 20.52  & 0.907  & 0.112
        & 22.60  & 0.829  & 0.082
        & 20.25  & 0.859   & 0.154
        & 23.92  & 0.947  & 0.063
        & 17.91  & 0.934   & 0.182
        &$\times$  &$\times$  &$\times$
        
        & 21.04  & 0.895   & 0.119 \\

        E + MotionGS \cite{MotionGS}
        & 22.64  & 0.876  & 0.166
        & 19.14   &  0.857 & 0.156
        & 15.39  & 0.823 & 0.217
        & 22.70  &  0.854  & 0.119
        & 16.99 & 0.902  & 0.115
        & 17.37  & 0.855  & 0.151
        
        &19.04  &0.861  &0.154  \\

        E + DynamicGS \cite{Dynamic3DGS}
        &    \textbf{36.32} &   \textbf{0.971}  &    \textbf{0.064}
        
        &  26.37  &  0.906  &  0.032 
        
        &  \textbf{28.44}  &  \textbf{0.939}  &  0.026
        &   31.81  & 0.942  &   0.050 
        
        &\textbf{25.36}   & 0.897  &\textbf{0.018}
        &$\times$   &$\times$  &$\times$ 
        
        &29.66   &0.931  &0.038  \\
        
        Ours
         &  33.11 &  0.964 &  0.015
         
        &   \textbf{28.32} &   \textbf{0.913} &  \textbf{0.029}
        
        &    26.07 &   0.937 &   \textbf{0.016}
        
        &  \textbf{33.08} &  \textbf{0.956} &  \textbf{0.019}
        & 24.76 & \textbf{0.917} & 0.024
        & \textbf{32.87} & \textbf{0.945}  & \textbf{0.016}
        
        & \textbf{29.70} & \textbf{0.939}  & \textbf{0.020}
        
        \\
        
        \bottomrule
    \end{tabular}
    }
        \caption{Quantitative  comparison of video reconstruction by different methods.}
    \label{3}
\end{table}

\begin{table}[t]
    \centering
    \renewcommand{\arraystretch}{1}
    \setlength{\tabcolsep}{3pt}
    \resizebox{1\textwidth}{!}{
    \begin{tabular}{l|*{6}{ccc|}cccc}
        \toprule
        & \multicolumn{3}{c|}{Grenade} 
        & \multicolumn{3}{c|}{Missile} 
        & \multicolumn{3}{c|}{Cola} 
        & \multicolumn{3}{c|}{Doll}
        & \multicolumn{3}{c|}{Chocolate} 
        & \multicolumn{3}{c|}{Rugby} 
        & \multicolumn{3}{c}{\textbf{Mean}}\\
        
        \cmidrule(lr){2-4}
        \cmidrule(lr){5-7} 
        \cmidrule(lr){8-10} 
        \cmidrule(lr){11-13} 
        \cmidrule(lr){14-16} 
        \cmidrule(lr){17-19}
        \cmidrule(lr){20-22} 
        
        & IoU$\uparrow$ & ATE$\downarrow$ & RMSE$\downarrow$ 
        & IoU$\uparrow$ & ATE$\downarrow$ & RMSE$\downarrow$ 
        & IoU$\uparrow$ & ATE$\downarrow$ & RMSE$\downarrow$ 
        & IoU$\uparrow$ & ATE$\downarrow$ & RMSE$\downarrow$ 
        & IoU$\uparrow$ & ATE$\downarrow$ & RMSE$\downarrow$ 
        & IoU$\uparrow$ & ATE$\downarrow$ & RMSE$\downarrow$                        
        & IoU$\uparrow$ & ATE$\downarrow$ & RMSE$\downarrow$ 

        \\
        
        \midrule

        CFGS \cite{CF}
        & 0.249 & 0.550 & 0.602
        & 0.203 & 0.129 & 0.218
        & 0.154 & 0.533 &  0.563
        & 0.310 & 0.358 & 0.459
        & 0.318  & 0.574  & 0.610
        & 0.094 & 0.562 & 0.604
        
        &0.221   &0.451 &0.509 \\

        E + 4DGS \cite{4DGS}
        & 0.359  & 0.661  & 0.703
        & 0.608  & 0.194  & 0.248
        & 0.354  & 0.630  & 0.661
        & 0.401  & 0.563  & 0.621
        & 0.287  & 0.668  & 0.687
        &$\times$   &$\times$  &$\times$
        
        &0.402   &0.543  &0.584 \\
        
        E + MotionGS \cite{MotionGS}
        & 0.288  & 0.474  & 0.505
        & 0.541   &  0.655 & 0.704
        & 0.423  & 0.511 & 0.551
        & 0.501  &  0.505  & 0.559
        & 0.549 & 0.501  & 0.522
        & 0.249 & 0.643  & 0.668

        &0.425   &0.548  &0.585
         \\

        E + DynamicGS \cite{Dynamic3DGS}
        & 0.548  & 0.227  &   0.269
        & 0.910  & 0.118  & 0.156  
        & 0.898  & 0.182  & 0.226
        & 0.663  & 0.311  & 0.363  
        &\textbf{0.995}   & \textbf{0.070}  &\textbf{0.090}
        &$\times$   &$\times$  &$\times$ 
        
        &0.803   &0.182  &0.221
      \\

        % E2GS \cite{e2gs}
        % &  &  & 
        % & -- & -- & --
        % & -- & -- & --
        % & -- & -- & --
        % & -- & -- & --
        % & -- & -- & --
        
        % &--   &--  &-- \\

        DeblurGS \cite{DeblurGS}
        & 0.693 & 0.217 & 0.253
        & 0.824 & 0.098 & 0.173
        & $\times$ & $\times$ & $\times$
        & 0.852 & 0.132 & 0.249
        & 0.947 & 0.186 & 0.233
        & 0.893 & 0.057 & 0.169
        
        &0.842   &0.138  &0.215\\

        ComoGS \cite{comogs}
        & 0.593 & 0.268 & 0.317
        & 0.732 & 0.179  & 0.177
        & 0.586 & 0.624 & 0.653
        & 0.649 & 0.160 & 0.273
        & 0.941 & 0.191 & 0.249
        & 0.880 & 0.070 & 0.185
        
        &0.730  &0.265  &0.309 \\

        Ours
        & \textbf{0.972} & \textbf{0.174} & \textbf{0.261}
        & \textbf{0.983} & \textbf{0.098} & \textbf{0.142}
        & \textbf{0.984} & \textbf{0.076} & \textbf{0.089}
        & \textbf{0.971} & \textbf{0.097} & \textbf{0.163}
        & 0.953 & 0.184 & 0.232
        & \textbf{0.946} & \textbf{0.029}  & \textbf{0.157}
        
        & \textbf{0.968} & \textbf{0.109}  & \textbf{0.174}
   \\
        
        \bottomrule
    \end{tabular}
    }
   \caption{Quantitative comparison of motion recovery by different methods on synthetic datasets.}
       \label{4}
\end{table}

\begin{table}[t]
    \centering
    \renewcommand{\arraystretch}{1}
    \setlength{\tabcolsep}{3pt}
    \resizebox{1\textwidth}{!}{
    \begin{tabular}{l|*{6}{ccc|}cccc}
        \toprule
        & \multicolumn{3}{c|}{Shark} 
        & \multicolumn{3}{c|}{Bear} 
        & \multicolumn{3}{c|}{Box1} 
        & \multicolumn{3}{c|}{Box2}
        & \multicolumn{3}{c|}{Box3} 
        & \multicolumn{3}{c|}{Doraemon} 
        & \multicolumn{3}{c}{\textbf{Mean}}\\
        
        \cmidrule(lr){2-4}
        \cmidrule(lr){5-7} 
        \cmidrule(lr){8-10} 
        \cmidrule(lr){11-13} 
        \cmidrule(lr){14-16} 
        \cmidrule(lr){17-19}
        \cmidrule(lr){20-22} 
        
        & IoU$\uparrow$ & ATE$\downarrow$ & RMSE$\downarrow$ 
        & IoU$\uparrow$ & ATE$\downarrow$ & RMSE$\downarrow$ 
        & IoU$\uparrow$ & ATE$\downarrow$ & RMSE$\downarrow$ 
        & IoU$\uparrow$ & ATE$\downarrow$ & RMSE$\downarrow$ 
        & IoU$\uparrow$ & ATE$\downarrow$ & RMSE$\downarrow$ 
        & IoU$\uparrow$ & ATE$\downarrow$ & RMSE$\downarrow$                        
        & IoU$\uparrow$ & ATE$\downarrow$ & RMSE$\downarrow$ 

        \\
        
        \midrule

        % CFGS \cite{CF}
        % &0.252  &0.425  &0.458
        % &0.199 &0.319 &0.364  
        % &0.079 &0.392 &0.425
        
        % &--   &--  &--
        % &--   &--  &--
        % &--   &--  &--
        
        % &--   &--  &-- \\

        E + 4DGS \cite{4DGS}
        & 0.141  & 0.505  & 0.556
        & $\times$ & $\times$ & $\times$
        & 0.144  & 0.553  & 0.592

        &0.076   &0.335  &0.374
        &0.190  &0.214  &0.305
        &0.202  &0.312  &0.448
        
        &0.151  &0.383  &0.566  \\

        E + DynamicGS \cite{Dynamic3DGS}
        & 0.679  & 0.094  &   0.155 
        & $\times$ & $\times$ & $\times$
        & 0.262  & 0.439  &  0.470 

        & 0.214 & 0.183 & 0.270
        & 0.378  & 0.152  &0.235
        &0.559   &0.146 &0.261
        
        &0.418   &0.203  &0.232 \\
    
        % E2GS \cite{e2gs}
        % & -- & -- & --
        % & -- & -- & --
        % & -- & -- & --
        % & -- & -- & --
        % & -- & -- & --
        % & -- & -- & --
        
        % &--   &--  &-- \\

        DeblurGS \cite{DeblurGS}
        & 0.834 & 0.033 & 0.142
        & 0.876 & \textbf{0.024} &\textbf{0.138}
        & 0.909 & \textbf{0.021} & 0.136
        & 0.195 & 0.236 & 0.356
        & 0.373 & 0.176 & 0.247
        & 0.497 & 0.208 & 0.317
        & 0.614 & 0.116 & 0.223 \\

        ComoGS \cite{comogs}
        & $\times$ & $\times$ & $\times$
        & 0.087 & 0.219 & 0.256
        & 0.053 & 0.211 & 0.242
        & 0.050 & 0.452 & 0.521
        & 0.292 & 0.201 & 0.317
        & 0.410 & 0.205 & 0.325
        
        &0.178   &0.257  &0.332 \\

        Ours
        & \textbf{0.862} & \textbf{0.027}  & \textbf{0.097}
        & \textbf{0.953} & 0.050  & 0.052
        & \textbf{0.931} & 0.066  & \textbf{0.094}
        & \textbf{0.664} & \textbf{0.057} & \textbf{0.060}
        &\textbf{0.822}   &\textbf{0.024}  & \textbf{0.028}
        & \textbf{0.628} & \textbf{0.075 }  & \textbf{0.167}
        
        & \textbf{0.810} & \textbf{0.050}  & \textbf{0.083}
   \\
        
        \bottomrule
    \end{tabular}
    }
   \caption{Quantitative comparison of motion recovery by different methods on real datasets.}
       \label{4}
\end{table}

\subsection{Comparison}
\noindent
\textbf{Comparison of video reconstruction.} 
Table.\ref{3} summarize the quantitative results of different methods for video reconstruction on real and synthetic data respectively. Our approach exhibits consistent performance. The introduced density control improves both visual and geometric fidelity, serving as a robust basis for recovering motion trajectories. \cite{CF} is limited by its depth inference pipeline. For modeling specific objects in constrained settings, monocular depth estimation faces an inherently underdetermined problem. This prevents the maintenance of 3D consistency across long sequences. By introducing physical or rigid constraints, \cite{Dynamic3DGS, MotionGS} achieve better results than the visual-only \cite{4DGS}.

\noindent
\textbf{Comparison of motion recovery.}  
Figure \ref{comparison} presents uniformly sampled frames from complete sequences, with quantitative results provided in Table \ref{4}. Among the evaluated methods, PMGS delivers strong performance across all three tracking metrics, indicating accurate estimation of both translational and rotational motion. In contrast, even when initialized with our well-trained 3D model, CFGS diverges significantly as its Gaussians are driven toward infinite spatial positions under purely photometric supervision in an attempt to fit image similarity. 4DGS and MotionGS exhibit noticeable trajectory fragmentation and visual artifacts, reflecting common failure modes of dynamic representations when confronted with large-scale motions. DynamicGS shows encouraging results but generalizes poorly to certain object instances. Although ComoGS and DeblurGS achieve reasonable motion trajectory estimation in most scenarios, they still fall short in complete motion deblurring. Collectively, these results underscore the importance of incorporating physical constraints and event-based cues for robust reconstruction of large blur motions.

\begin{figure}[th]
    \centering
    \includegraphics[width=\textwidth]{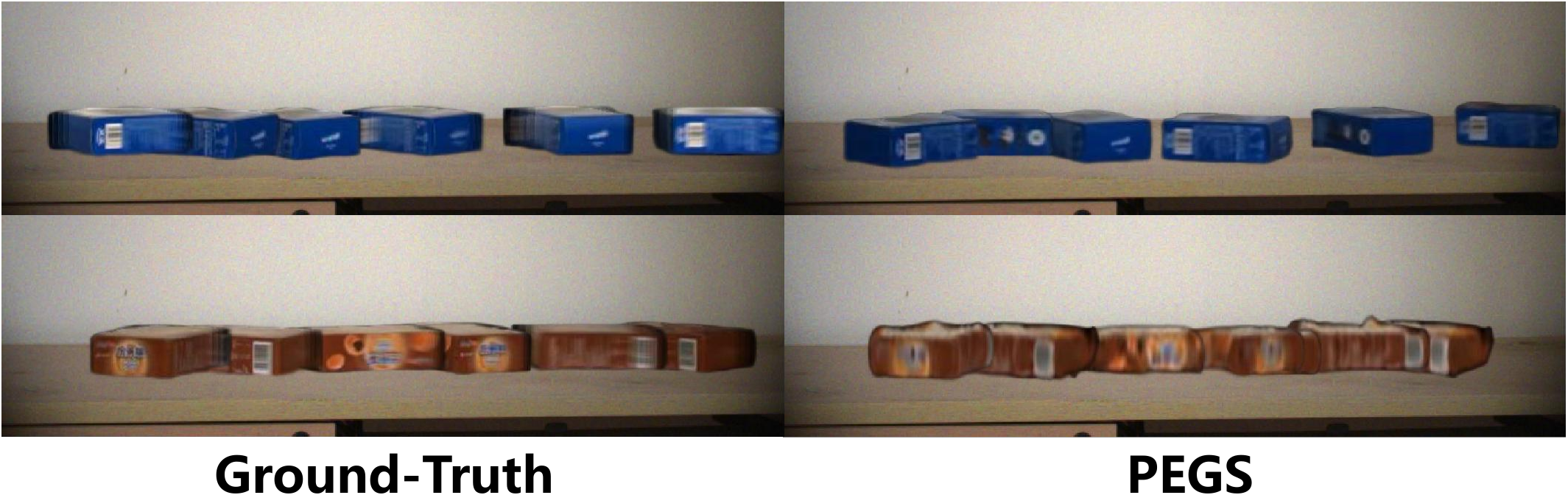} 
    \caption{ While PEGS exhibits minor differences in color rendering, it accurately recovers the motion trajectory under fast friction.}
    \label{comparison}
\end{figure}
\label{sec:experiments}

\begin{table}[t]
    \centering
    \renewcommand{\arraystretch}{1}
    \setlength{\tabcolsep}{4pt}

    \label{ablation}
    \resizebox{\textwidth}{!}{
    \begin{tabular}{l | c c c | c c c c}
        \toprule
        & \multicolumn{3}{c|}{\textbf{Video reconstruction}} & \multicolumn{4}{c}{\textbf{Motion recovery}} \\
        \cmidrule(lr){2-4} \cmidrule(lr){5-8}
        & PSNR$\uparrow$ & SSIM$\uparrow$ & LPIPS$\downarrow$ & IoU$\uparrow$ & ATE$\downarrow$ & RMSE$\downarrow$ & Times$\downarrow$ \\
        \midrule
        
        DeblurGS  
        & 25.01  & 0.895  & 0.073   
        &  0.728 & 0.127  & 0.219 
        &  23 hours\\
        
        w/o density control     
        & 27.47  & 0.918  &  0.050  
        & 0.863  & 0.133  &  0.184
        & 57 mins\\
        
        w/o $\mathcal{\bm{L}}_{Acc}$ \& MSA      
        & 26.35  & 0.912  & 0.054   
        & 0.847  & 0.140  &  0.211
        &  72 mins\\

        w/o $\mathcal{\bm{L}}_{Event}$     
        & 28.14  &  0.924 & 0.048   
        & 0.926 & 0.103  &  0.179
        & 51 mins\\
        
        w/o $\mathcal{\bm{L}}_{KF}$
        & 29.21 &  0.939 & 0.041   
        & 0.832  & 0.096  &  0.152
        & \textbf{46 mins}\\

        PEGS (Full) 
& \textbf{29.70  }& \textbf{0.950 }&  \textbf{0.032} 
        & \textbf{0.889}  &  \textbf{0.080} &  \textbf{0.129}
        & 56 mins\\

        \bottomrule
    \end{tabular}
    }
        \caption{The full model achieved the best performance, with a substantially lower training time than DeblurGS.}
        \label{ablatable}
\end{table}

\subsection{Ablation Study}

% \noindent
\textbf{Effectiveness of point density control.} As shown in Table \ref{ablatable}, the performance drop of Model 1 suggest that poor reconstruction quality adversely affects tracking accuracy. Inaccurate geometric and appearance representations (floating Gaussians and artifacts) may lead to target localization deviations, resulting in accumulated errors over time.

\noindent
\textbf{Effectiveness of physical constraints.}
In Model 2, the $\mathcal{\bm{L}}_{Acc}$ and MSA strategy are excluded, while retaining only photometric supervision and a fixed training paradigm. Trajectory accuracy declines significantly, with AME increasing 0.06 and RMSE increasing by 0.82. Additionally, the removal of the MSA strategy substantially increases the optimization cost and introduces potential risks of training oscillation.

\noindent
\textbf{Effectiveness of event enhancement.} The removal of $\mathcal{\bm{L}}_{Event}$ led to degradation in both reconstruction and motion recovery (1.56 dB drop in PSNR and a 0.05 decline in RMSE), which demonstrates the high temporal resolution supervision provided by event streams compensates for motion cues lost by RGB cameras when capturing large motions.

\noindent
\textbf{Effectiveness of Kalman fusion.} Though removing the $\mathcal{\bm{L}}_{KF}$ regularization slightly reduces computational costs, it leads to a decrease in stability, quantified as a 0.49 dB drop in PSNR and a 0.016 increase in ATE. In more challenging scenarios, such as when external disturbances are present or the captured event stream data contains pronounced noise (e.g., in low-light conditions), the role of this strategy in ensuring system robustness would become more critical.

\section{Conclusion}
\label{sec:conclusion}

In this paper, we present PEGS for the reconstruction of large spatiotemporal motion. By combining a triple-level supervision scheme featuring acceleration constraints, event stream enhancement, and Kalman fusion with MSA strategy, we effectively tackle physical inaccuracy and motion blur while ensuring robust convergence. The explicit and physically consistent motion representation offered by PEGS establishes a foundation for dynamic scene understanding and simulation, showing promising potential for exploring real-world 3D physical interactions. Our future investigations will be directed towards addressing demanding scenarios, exemplified by systems with variable force fields, targets at extreme scales, and low-quality image acquisition.

\bibliographystyle{plain} 
\bibliography{main}

@String(CVPR= {IEEE Conf. Comput. Vis. Pattern Recog.})

@String(ICCV= {Int. Conf. Comput. Vis.})

@String(CVPRW= {IEEE Conf. Comput. Vis. Pattern Recog. Worksh.})

@String(CVPR  = {CVPR})

@String(ICCV  = {ICCV})

@String(CVPRW= {CVPRW})

@article{3DGS,
  title={3d gaussian splatting for real-time radiance field rendering.},
  author={Kerbl, Bernhard and Kopanas, Georgios and Leimk{\"u}hler, Thomas and Drettakis, George},
  journal={ACM Trans. Graph.},
  volume={42},
  number={4},
  pages={139--1},
  year={2023}
}

@inproceedings{DeformGS,
  title={Deformable 3d gaussians for high-fidelity monocular dynamic scene reconstruction},
  author={Yang, Ziyi and Gao, Xinyu and Zhou, Wen and Jiao, Shaohui and Zhang, Yuqing and Jin, Xiaogang},
  booktitle={Proceedings of the IEEE/CVF conference on computer vision and pattern recognition},
  pages={20331--20341},
  year={2024}
}

@inproceedings{Dynamic3DGS,
  title={Dynamic 3d gaussians: Tracking by persistent dynamic view synthesis},
  author={Luiten, Jonathon and Kopanas, Georgios and Leibe, Bastian and Ramanan, Deva},
  booktitle={2024 International Conference on 3D Vision (3DV)},
  pages={800--809},
  year={2024},
  organization={IEEE}
}

@inproceedings{4DGS,
  title={4d gaussian splatting for real-time dynamic scene rendering},
  author={Wu, Guanjun and Yi, Taoran and Fang, Jiemin and Xie, Lingxi and Zhang, Xiaopeng and Wei, Wei and Liu, Wenyu and Tian, Qi and Wang, Xinggang},
  booktitle={Proceedings of the IEEE/CVF conference on computer vision and pattern recognition},
  pages={20310--20320},
  year={2024}
}

@article{Dream4DGS,
  title={Dreamgaussian4d: Generative 4d gaussian splatting},
  author={Ren, Jiawei and Pan, Liang and Tang, Jiaxiang and Zhang, Chi and Cao, Ang and Zeng, Gang and Liu, Ziwei},
  journal={arXiv preprint arXiv:2312.17142},
  year={2023}
}

@inproceedings{FSGS,
  title={Fsgs: Real-time few-shot view synthesis using gaussian splatting},
  author={Zhu, Zehao and Fan, Zhiwen and Jiang, Yifan and Wang, Zhangyang},
  booktitle={European conference on computer vision},
  pages={145--163},
  year={2024},
  organization={Springer}
}

@inproceedings{CF,
  title={Colmap-free 3d gaussian splatting},
  author={Fu, Yang and Liu, Sifei and Kulkarni, Amey and Kautz, Jan and Efros, Alexei A and Wang, Xiaolong},
  booktitle={Proceedings of the IEEE/CVF Conference on Computer Vision and Pattern Recognition},
  pages={20796--20805},
  year={2024}
}

@inproceedings{2DGS,
  title={2d gaussian splatting for geometrically accurate radiance fields},
  author={Huang, Binbin and Yu, Zehao and Chen, Anpei and Geiger, Andreas and Gao, Shenghua},
  booktitle={ACM SIGGRAPH 2024 conference papers},
  pages={1--11},
  year={2024}
}

@article{Instantsplat,
  title={Instantsplat: Unbounded sparse-view pose-free gaussian splatting in 40 seconds},
  author={Fan, Zhiwen and Cong, Wenyan and Wen, Kairun and Wang, Kevin and Zhang, Jian and Ding, Xinghao and Xu, Danfei and Ivanovic, Boris and Pavone, Marco and Pavlakos, Georgios and others},
  journal={arXiv preprint arXiv:2403.20309},
  volume={2},
  number={3},
  pages={4},
  year={2024}
}

@inproceedings{Endo,
  title={Endo-4dgs: Endoscopic monocular scene reconstruction with 4d gaussian splatting},
  author={Huang, Yiming and Cui, Beilei and Bai, Long and Guo, Ziqi and Xu, Mengya and Islam, Mobarakol and Ren, Hongliang},
  booktitle={International Conference on Medical Image Computing and Computer-Assisted Intervention},
  pages={197--207},
  year={2024},
  organization={Springer}
}

@inproceedings{Spring,
  title={Reconstruction and simulation of elastic objects with spring-mass 3d gaussians},
  author={Zhong, Licheng and Yu, Hong-Xing and Wu, Jiajun and Li, Yunzhu},
  booktitle={European Conference on Computer Vision},
  pages={407--423},
  year={2024},
  organization={Springer}
}

@inproceedings{PEGAUS,
  title={Pegasus: Physically enhanced gaussian splatting simulation system for 6dof object pose dataset generation},
  author={Meyer, Lukas and Erich, Floris and Yoshiyasu, Yusuke and Stamminger, Marc and Ando, Noriaki and Domae, Yukiyasu},
  booktitle={2024 IEEE/RSJ International Conference on Intelligent Robots and Systems (IROS)},
  pages={10710--10715},
  year={2024},
  organization={IEEE}
}

@article{Dreamscene4D,
  title={Dreamscene4d: Dynamic multi-object scene generation from monocular videos},
  author={Chu, Wen-Hsuan and Ke, Lei and Fragkiadaki, Katerina},
  journal={arXiv preprint arXiv:2405.02280},
  year={2024}
}

@inproceedings{DynaNeRF2,
  title={D-nerf: Neural radiance fields for dynamic scenes},
  author={Pumarola, Albert and Corona, Enric and Pons-Moll, Gerard and Moreno-Noguer, Francesc},
  booktitle={Proceedings of the IEEE/CVF conference on computer vision and pattern recognition},
  pages={10318--10327},
  year={2021}
}

@article{DynaNeRF3,
  title={Hypernerf: A higher-dimensional representation for topologically varying neural radiance fields},
  author={Park, Keunhong and Sinha, Utkarsh and Hedman, Peter and Barron, Jonathan T and Bouaziz, Sofien and Goldman, Dan B and Martin-Brualla, Ricardo and Seitz, Steven M},
  journal={arXiv preprint arXiv:2106.13228},
  year={2021}
}

@article{NeRFPlayer,
  title={Nerfplayer: A streamable dynamic scene representation with decomposed neural radiance fields},
  author={Song, Liangchen and Chen, Anpei and Li, Zhong and Chen, Zhang and Chen, Lele and Yuan, Junsong and Xu, Yi and Geiger, Andreas},
  journal={IEEE Transactions on Visualization and Computer Graphics},
  volume={29},
  number={5},
  pages={2732--2742},
  year={2023},
  publisher={IEEE}
}

@inproceedings{DyNeRF,
  title={Neural 3d video synthesis from multi-view video},
  author={Li, Tianye and Slavcheva, Mira and Zollhoefer, Michael and Green, Simon and Lassner, Christoph and Kim, Changil and Schmidt, Tanner and Lovegrove, Steven and Goesele, Michael and Newcombe, Richard and others},
  booktitle={Proceedings of the IEEE/CVF conference on computer vision and pattern recognition},
  pages={5521--5531},
  year={2022}
}

@inproceedings{SAM,
  title={Segment anything},
  author={Kirillov, Alexander and Mintun, Eric and Ravi, Nikhila and Mao, Hanzi and Rolland, Chloe and Gustafson, Laura and Xiao, Tete and Whitehead, Spencer and Berg, Alexander C and Lo, Wan-Yen and others},
  booktitle={Proceedings of the IEEE/CVF international conference on computer vision},
  pages={4015--4026},
  year={2023}
}

@inproceedings{SfM,
  title={Structure-from-motion revisited},
  author={Schonberger, Johannes L and Frahm, Jan-Michael},
  booktitle={Proceedings of the IEEE conference on computer vision and pattern recognition},
  pages={4104--4113},
  year={2016}
}

@inproceedings{Tracking,
  title={Tracking by 3d model estimation of unknown objects in videos},
  author={Rozumnyi, Denys and Matas, Ji{\v{r}}{\'\i} and Pollefeys, Marc and Ferrari, Vittorio and Oswald, Martin R},
  booktitle={Proceedings of the IEEE/CVF International Conference on Computer Vision},
  pages={14086--14096},
  year={2023}
}

@inproceedings{DPT,
  title={Vision transformers for dense prediction},
  author={Ranftl, Ren{\'e} and Bochkovskiy, Alexey and Koltun, Vladlen},
  booktitle={Proceedings of the IEEE/CVF international conference on computer vision},
  pages={12179--12188},
  year={2021}
}

@misc{Blender,
  author = {{Blender Foundation}},
  title = {{Blender} --- a 3D Modelling and Rendering Package},
  year = {2018},
  howpublished = {Blender Foundation, Amsterdam},
  url = {https://www.blender.org},
}

@article{SSIM,
  title={Image quality assessment: from error visibility to structural similarity},
  author={Wang, Zhou and Bovik, Alan C and Sheikh, Hamid R and Simoncelli, Eero P},
  journal={IEEE transactions on image processing},
  volume={13},
  number={4},
  pages={600--612},
  year={2004},
  publisher={IEEE}
}

@inproceedings{PSNR,
  title={The unreasonable effectiveness of deep features as a perceptual metric},
  author={Zhang, Richard and Isola, Phillip and Efros, Alexei A and Shechtman, Eli and Wang, Oliver},
  booktitle={Proceedings of the IEEE conference on computer vision and pattern recognition},
  pages={586--595},
  year={2018}
}

@inproceedings{IoU,
  title={Unitbox: An advanced object detection network},
  author={Yu, Jiahui and Jiang, Yuning and Wang, Zhangyang and Cao, Zhimin and Huang, Thomas},
  booktitle={Proceedings of the 24th ACM international conference on Multimedia},
  pages={516--520},
  year={2016}
}

@article{Kalman,
  title={An introduction to the Kalman filter},
  author={Welch, Greg and Bishop, Gary and others},
  year={1995},
  publisher={Chapel Hill, NC, USA}
}

@article{MotionAware,
  title={Motion-Aware 3D Gaussian Splatting for Efficient Dynamic Scene Reconstruction},
  author={Zhiyang Guo and Wen-gang Zhou and Li Li and Min Wang and Houqiang Li},
  journal={IEEE Transactions on Circuits and Systems for Video Technology},
  year={2024},
  volume={35},
  pages={3119-3133},
  url={https://api.semanticscholar.org/CorpusID:268512916}
}

@article{MotionGS,
  title={MotionGS: Exploring Explicit Motion Guidance for Deformable 3D Gaussian Splatting},
  author={Ruijie Zhu and Yanzhe Liang and Hanzhi Chang and Jiacheng Deng and Jiahao Lu and Wenfei Yang and Tianzhu Zhang and Yongdong Zhang},
  journal={ArXiv},
  year={2024},
  volume={abs/2410.07707},
  url={https://api.semanticscholar.org/CorpusID:273234208}
}

@article{particlegs,
  title={ParticleGS: Particle-Based Dynamics Modeling of 3D Gaussians for Prior-free Motion Extrapolation},
  author={Jinsheng Quan and Chunshi Wang and Yawei Luo},
  journal={ArXiv},
  year={2025},
  volume={abs/2505.20270},
  url={https://api.semanticscholar.org/CorpusID:278911262}
}

@article{SAGS,
  title={SA-GS: Semantic-Aware Gaussian Splatting for Large Scene Reconstruction with Geometry Constrain},
  author={Butian Xiong and Xiaoyu Ye and Tze Ho Elden Tse and Kai Han and Shuguang Cui and Zhen Li},
  journal={ArXiv},
  year={2024},
  volume={abs/2405.16923},
  url={https://api.semanticscholar.org/CorpusID:270063485}
}

@article{S4d,
  title={S4D: Streaming 4D Real-World Reconstruction with Gaussians and 3D Control Points},
  author={Bing He and Yunuo Chen and Guo Lu and Li Song and Wenjun Zhang},
  journal={ArXiv},
  year={2024},
  volume={abs/2408.13036},
  url={https://api.semanticscholar.org/CorpusID:271946844}
}

@article{comogs,
  title={CoMoGaussian: Continuous Motion-Aware Gaussian Splatting from Motion-Blurred Images},
  author={Lee, Jungho and Kim, Donghyeong and Lee, Dogyoon and Cho, Suhwan and Lee, Minhyeok and Lee, Wonjoon and Kim, Taeoh and Wee, Dongyoon and Lee, Sangyoun},
  journal={arXiv preprint arXiv:2503.05332},
  year={2025}
}

@article{pybullet,
  title={PyBullet quickstart guide},
  author={Coumans, Erwin and Bai, Yunfei},
  journal={ed: PyBullet Quickstart Guide. https://docs. google. com/document/u/1/d},
  year={2021}
}

@article{v2e,
  title={v2e: From Video Frames to Realistic DVS Events},
  author={Yuhuang Hu and Shih-Chii Liu and Tobi Delbruck},
  journal={2021 IEEE/CVF Conference on Computer Vision and Pattern Recognition Workshops (CVPRW)},
  year={2021},
  pages={1312-1321},
  url={https://api.semanticscholar.org/CorpusID:235651771}
}

@article{e2nerf,
  title={E2NeRF: Event Enhanced Neural Radiance Fields from Blurry Images},
  author={Yunshan Qi and Lin Zhu and Yu Zhang and Jia Li},
  journal={2023 IEEE/CVF International Conference on Computer Vision (ICCV)},
  year={2023},
  pages={13208-13218},
  url={https://api.semanticscholar.org/CorpusID:265427501}
}

@article{evdeblurnerf,
  title={Mitigating Motion Blur in Neural Radiance Fields with Events and Frames},
  author={Marco Cannici and Davide Scaramuzza},
  journal={2024 IEEE/CVF Conference on Computer Vision and Pattern Recognition (CVPR)},
  year={2024},
  pages={9286-9296},
  url={https://api.semanticscholar.org/CorpusID:268793644}
}

@article{denerf,
  title={DE-NeRF: DEcoupled Neural Radiance Fields for View-Consistent Appearance Editing and High-Frequency Environmental Relighting},
  author={Tong Wu and Jiali Sun and Yu-Kun Lai and Lin Gao},
  journal={ACM SIGGRAPH 2023 Conference Proceedings},
  year={2023},
  url={https://api.semanticscholar.org/CorpusID:259975832}
}

@article{e2gs,
  title={E2GS: Event Enhanced Gaussian Splatting},
  author={Hiroyuki Deguchi and Mana Masuda and Takuya Nakabayashi and Hideo Saito},
  journal={ArXiv},
  year={2024},
  volume={abs/2406.14978},
  url={https://api.semanticscholar.org/CorpusID:270688286}
}

@article{EaDeblur,
  title={EaDeblur-GS: Event assisted 3D Deblur Reconstruction with Gaussian Splatting},
  author={Yuchen Weng and Zhengwen Shen and Ruofan Chen and Qi Wang and Jun Wang},
  journal={ArXiv},
  year={2024},
  volume={abs/2407.13520},
  url={https://api.semanticscholar.org/CorpusID:271270603}
}

@article{e4dgs,
  title={E-4DGS: High-Fidelity Dynamic Reconstruction from the Multi-view Event Cameras},
  author={Chaoran Feng and Zhenyu Tang and Wangbo Yu and Yatian Pang and Yian Zhao and Jianbin Zhao and Li Yuan and Yonghong Tian},
  journal={ArXiv},
  year={2025},
  volume={abs/2508.09912},
  url={https://api.semanticscholar.org/CorpusID:280641698}
}

@inproceedings{Eventboost,
  title={Event-boosted Deformable 3D Gaussians for Dynamic Scene Reconstruction},
  author={Wenhao Xu and Wenming Weng and Yueyi Zhang and Ruikang Xu and Zhiwei Xiong},
  year={2024},
  url={https://api.semanticscholar.org/CorpusID:274234776}
}

@article{degs,
  title={DEGS: Deformable Event-based 3D Gaussian Splatting from RGB and Event Stream},
  author={Li, Jia and Wang, Jiaxu and He, Junhao and Sun, Mingyuan and Xu, Renjing and Zhang, Qiang and Cao, Jiahang and Zhang, Ziyi and Gu, Yi and SUN, Jingkai},
  year={2025}
}

@article{event3dgs,
  title={Event3dgs: Event-based 3d gaussian splatting for high-speed robot egomotion},
  author={Xiong, Tianyi and Wu, Jiayi and He, Botao and Fermuller, Cornelia and Aloimonos, Yiannis and Huang, Heng and Metzler, Christopher A},
  journal={arXiv preprint arXiv:2406.02972},
  year={2024}
}

@article{Darkgs,
  title={Dark-EvGS: Event Camera as an Eye for Radiance Field in the Dark},
  author={Wu, Jingqian and Duan, Peiqi and Wang, Zongqiang and Wang, Changwei and Shi, Boxin and Lam, Edmund Y},
  journal={arXiv preprint arXiv:2507.11931},
  year={2025}
}

@inproceedings{EDI,
  title={Bringing a blurry frame alive at high frame-rate with an event camera},
  author={Pan, Liyuan and Scheerlinck, Cedric and Yu, Xin and Hartley, Richard and Liu, Miaomiao and Dai, Yuchao},
  booktitle={Proceedings of the IEEE/CVF conference on computer vision and pattern recognition},
  pages={6820--6829},
  year={2019}
}

@article{evagaussians,
  title={Evagaussians: Event stream assisted gaussian splatting from blurry images},
  author={Yu, Wangbo and Feng, Chaoran and Tang, Jiye and Yang, Jiashu and Tang, Zhenyu and Jia, Xu and Yang, Yuchao and Yuan, Li and Tian, Yonghong},
  journal={arXiv preprint arXiv:2405.20224},
  year={2024}
}

@inproceedings{eraft,
  title={E-raft: Dense optical flow from event cameras},
  author={Gehrig, Mathias and Millh{\"a}usler, Mario and Gehrig, Daniel and Scaramuzza, Davide},
  booktitle={2021 International Conference on 3D Vision (3DV)},
  pages={197--206},
  year={2021},
  organization={IEEE}
}

@article{DeblurGS,
  title={Deblur-GS: 3D Gaussian Splatting from Camera Motion Blurred Images},
  author={Wenbo Chen and Ligang Liu},
  journal={Proceedings of the ACM on Computer Graphics and Interactive Techniques},
  year={2024},
  volume={7},
  pages={1 - 15},
  url={https://api.semanticscholar.org/CorpusID:269825809}
}

@inproceedings{e-boost,
  title={Event-boosted Deformable 3D Gaussians for Dynamic Scene Reconstruction},
  author={Xu, Wenhao and Weng, Wenming and Zhang, Yueyi and Xu, Ruikang and Xiong, Zhiwei},
  booktitle={Proceedings of the IEEE/CVF International Conference on Computer Vision},
  pages={28334--28343},
  year={2025}
}

@article{pmgs,
  title={PMGS: Reconstruction of Projectile Motion across Large Spatiotemporal Spans via 3D Gaussian Splatting},
  author={Yijun Xu and Jingrui Zhang and Yuhan Chen and Dingwen Wang and Lei Yu and Chu He},
  journal={ArXiv},
  year={2025},
  volume={abs/2508.02660},
  url={https://api.semanticscholar.org/CorpusID:280422270}
}

\end{document}